\documentclass{article}

\usepackage{microtype}
\usepackage{graphicx}
\usepackage{subfigure}
\usepackage{booktabs} 
\usepackage{enumitem}
\usepackage{multirow}
\usepackage{graphicx} 

\usepackage{hyperref}
\usepackage{url}
\usepackage{graphicx}
\usepackage{wrapfig}
\usepackage{algpseudocode}
\usepackage{amsmath}
\usepackage{amssymb}
\usepackage{xcolor}
\usepackage{tcolorbox}
\usepackage{listings}
\usepackage{pgfplots}
\usepackage{caption}
\usepackage{booktabs}
\usepackage[table]{xcolor}
\usepackage{float}      
\usepackage{placeins}  
\usepackage{colortbl}
\usepackage{stfloats}
\tcbuselibrary{skins,breakable}
\usepackage{fvextra}
\newtcolorbox{promptbox}[1]{
  colback=black!5,      
  colframe=black!75,    
  fonttitle=\bfseries,
  title=#1,
  arc=2mm,              
  boxsep=5pt,
}

\newcommand{\qwenlarge}{{\textsc{Qwen-2.5-7B-\allowbreak Instruct}}}
\newcommand{\qwenthree}{{\textsc{Qwen3-8B}}}
\newcommand{\qwensmall}{{\textsc{Qwen-2.5-3B-\allowbreak Instruct}}}
\newcommand{\llama}{\textsc{LLaMA3.1-8B-\allowbreak Instruct}}

\usepgfplotslibrary{groupplots}
\pgfplotsset{compat=1.18}
\definecolor{darkblue}{RGB}{26, 13, 171}
\definecolor{darkgreen}{RGB}{0, 128, 0}
\definecolor{darkred}{RGB}{139, 0, 0}
\usepackage[table]{xcolor}
\algrenewcommand\algorithmicrequire{\textbf{Input:}}
\algrenewcommand\algorithmicensure{\textbf{Output:}}
\algrenewcommand\algorithmiccomment[1]{\hfill\textcolor{darkgreen}{\(\triangleright\) \textit{#1}}}

\newcommand{\methodname}{DiSC}

\usepackage{hyperref}

\usepackage[preprint]{icml2026}

\usepackage{amsmath}
\usepackage{amssymb}
\usepackage{mathtools}
\usepackage{amsthm}

\usepackage[capitalize,noabbrev]{cleveref}

\theoremstyle{plain}

\theoremstyle{definition}

\theoremstyle{remark}

\usepackage[textsize=tiny]{todonotes}

\icmltitlerunning{Context Distillation Retains Post-Training Capabilities in Continually Trained LMs}

\begin{document}

\twocolumn[
  \icmltitle{Updating Parametric Knowledge with Context Distillation \\ Retains Post-Training Capabilities}





  \icmlsetsymbol{equal}{*}

  \begin{icmlauthorlist}
    \icmlauthor{Shankar Padmanabhan}{yyy}
    \icmlauthor{Mustafa Omer Gul}{yyy}
    \icmlauthor{Tanya Goyal}{yyy}
  \end{icmlauthorlist}

  \icmlaffiliation{yyy}{Department of Computer Science, Cornell University}

  \icmlcorrespondingauthor{Shankar Padmanabhan}{sp2583@cornell.edu}

  \icmlkeywords{Machine Learning, ICML}

  \vskip 0.3in
]


%



\printAffiliationsAndNotice{}  

\setcounter{footnote}{1}


\begin{abstract}
Post-training endows pretrained LLMs with a variety of desirable skills, including instruction-following, reasoning, and others. However, these post-trained LLMs only encode knowledge up to a cut-off date, necessitating continual adaptation. Unfortunately, existing solutions cannot simultaneously learn new knowledge from an adaptation document corpora and mitigate the forgetting of earlier learned capabilities.
To address this, we introduce Distillation via Split Contexts (DiSC), a simple context-distillation based approach for continual knowledge adaptation. \methodname~derives student and teacher distributions by conditioning on distinct segments of the training example and minimizes the KL divergence between the shared tokens. This allows us to efficiently apply context-distillation without requiring explicit generation steps during training. 
We run experiments on four post-trained models and two adaptation domains. Compared to prior finetuning and distillation methods for continual adaptation, DiSC consistently reports the best trade-off between learning new knowledge and mitigating forgetting of previously learned skills like instruction-following, reasoning, and factual knowledge.\footnote{Code is publicly shared at \url{https://github.com/shankarp8/distillation-retains-capabilities}}

\end{abstract}

\section{Introduction}
Frontier large language models (LLMs) owe much of their general-purpose problem solving abilities, such as reasoning, coding, and instruction-following, to extensive post-training and alignment procedures. 
However, these LLMs remain static reflections of the data they were exposed to in their training phase while many real-world applications require adapting these LLMs after initial deployment.

We can broadly classify continual adaptation goals into two categories: adapting a model's \emph{capabilities} (e.g., to learn a new task or modify behavior), and learning new \emph{knowledge} (e.g., data from after pretraining, internal company information, etc.). Although standard finetuning for these new goals often results in severe catastrophic forgetting in both settings, recent work has shown promising results for capability adaptation when the task domains are amenable to RL-training.  In particular, \citet{shenfeld2025rlsrazoronlinereinforcement} and~\citet{chen2025retaining} show that RL-based updates can mitigate catastrophic forgetting as they implicitly favor solutions that minimize the KL divergence from the initial policy. 




However, continually adapting models to new knowledge domains, such as to a new document corpora, remains challenging. These raw documents lack verifiable reward signals, making it difficult to apply RL solutions. As a result, the dominant paradigm in this regime is supervised continual learning, which attempt to address catastrophic forgetting by constraining parameters, gradients, or data mixtures during fine-tuning \citep{ouyang2022training, deepseek,gu-etal-2024-model}. However, these modifications still rely on next-token prediction as their primary supervision signal, treating the preservation of prior behavior as an \emph{auxiliary} objective. Empirically, these approaches fail to reliably mitigate forgetting in continual adaptation.


In this work, we propose \textbf{Distillation via Split Contexts (DiSC)}, a context distillation method for continual knowledge adaptation of post-trained LLMs. Rather than treating behavior preservation as an auxiliary regularizer,
\methodname~uses the KL divergence as its primary objective to achieve behavior preservation. 
Particularly, DiSC trains the model to match predictions of a teacher model derived from the same model policy, but conditioned on a richer context that contains the new knowledge we want the LLM to learn. Thus, \methodname~effectively minimizes behavioral divergence while incorporating new knowledge in the trained student model.



To evaluate DiSC, we run experiments on four post-trained LLMs: \qwenlarge, \qwensmall~, \llama, and \qwenthree. We train for domain adaptation on two domains, KUP \cite{li2025memorization}, a news-style dataset containing synthetic updates to world knowledge, and a biomedical dataset BioASQ \cite{krithara2023bioasq}. We compare DiSC against prior fine-tuning techniques (e.g., with on-policy data \cite{chen2025retaining}, token-adaptive loss \citep{talr}, etc.) and knowledge distillation approaches proposed for continual adaptation \cite{padmanabhan2023propagating}. In addition to domain adaptation, we measure catastrophic forgetting for a suite of 7 standard tasks, including Big-Bench Hard \citep{bbh}, GPQA \citep{gpqa}, IFEval \citep{ifeval}, MATH-Hard \citep{hendrycks2021math}, MuSR \citep{musr}, MMLU-Pro \citep{mmlu-pro}, and the coding benchmark HumanEval \citep{chen2021codex}. 
 
\textbf{Finding 1: DiSC mitigates catastrophic forgetting during continual knowledge adaptation.} Our results show that DiSC outperforms prior methods on domain adaptation, while incurring negligible degradation on instruction-following, reasoning and other prior capabilities. For instance, when adapting \qwenlarge~to a document-level corpus, finetuning degrades instruction following by nearly 15 points and math reasoning by over 25 points. In contrast, DiSC limits degradation on both domains to less than 5 points, while achieving \emph{superior} domain gains.  These results establish context distillation-based supervision as an effective and performant technique for continual knowledge updating in post-trained LLMs.



\textbf{Finding 2: Existing finetuning based methods fail to alleviate catastrophic forgetting.} In particular, we observe a consistent and robust trade-off between capability preservation and domain adaptation that these methods fail to improve. For instance, on-policy finetuning \citep{chen2025retaining}, TALR \citep{talr}, and KL regularization result in roughly the same degradation on IFEval, MATH, and HumanEval performance as finetuning, across settings. We found that while finetuning with LoRA often reduces forgetting, it comes at the expense of in-domain performance. 

\textbf{Finding 3: ``Post-training skills'' suffer the largest degradation after finetuning for continual adaptation.} To understand task degradation after finetuning, we loosely classify tasks into pre-training and post-training skills depending on whether the improvement between the pre- and post-training checkpoints was minimal or significant respectively. Interestingly, we find that post-training skills, such as instruction-following, math reasoning and coding, show the largest degradation after adaptation with finetuning (e.g. $\sim$8-15 points drop in IFEval across models and finetuning baselines on KUP). On the other hand, pretraining skills that did not see improvement with post-training, such as BBH, GPQA, MMLU and MuSR, report negligible degradation. In fact, the improvement in the task performances between pre- and post-training is strongly correlated ($\approx 0.9$) to its subsequent drop after regular finetuning. These results provide strong evidence that finetuning reverts the post-trained model back towards its pre-trained behavior. 

\section{Continual Knowledge Adaptation}

\subsection{Problem Definition}
\label{sec:prb_def}
We study the problem of continual knowledge adaptation, where the task is to adapt an LLM to internalize new knowledge while mitigating the catastrophic forgetting of existing capabilities. In order to emulate realistic update scenarios, we make two design decisions when formalizing our setting: 

\textbf{First, we assume that the new knowledge is provided in the form of multi-sentence or multi-paragraph documents}. These documents might reflect updates to world knowledge after the training cut-off date (e.g. news articles), data unseen during prior training or even specialized data for a target domain such as medicine or law. This differs from earlier works in knowledge editing and updating that unrealistically assume that knowledge changes are easily accessible in the form of ``factoids'' or key-value tuples~\citep{meng2022rome, mitchell2022mend, meng2023memit, padmanabhan2023propagating, Dai2021KnowledgeNI}.  

\textbf{Second, we focus on adapting post-trained LLMs.} Prior research has traditionally explored adaptation of ``raw'' documents, as in our setting, for pre-trained models \citep{yang2024syntheticcontinuedpretraining,li2025memorization,su2024conflictbank}.
However, post-training significantly improves upon the pretrained LLM on several key dimensions, including instruction-following, reasoning, alignment with human preferences, among others. It is infeasible to adapt the pre-trained model on new knowledge and then repeat this complex and expensive post-training pipeline. Therefore, in this paper, we focus on continual knowledge adaptation of post-trained LLMs, where, in addition to adaptation of new knowledge, both pretraining \textit{and} post-training capabilities must be preserved.


\paragraph{Notation}
Formally, let $M_{\text{post}}$ be a post-trained language model obtained after alignment training of a pretrained model $M_{\text{base}}$. Let $\mathcal{D}=\{d^{(1)}, \ldots, d^{(m)}\}$ be a corpus of documents reflecting the new knowledge. Our task is to design an adaptation procedure that is given $M_{\text{post}}$ and $\mathcal{D}$ as input and produces an updated model $M_{\text{new}}$ that internalizes the new knowledge in $\mathcal{D}$ while preserving the general capabilities of $M_{\text{post}}$. The task setup does not specify any details about how the new knowledge in $\mathcal{D}$ will be tested or what the set of general capability tasks will be during training.

The trained models' performance is measured on two different benchmark suites, reflecting the dual goals of the problem setting: 1) \textbf{Knowledge Adaptation}: Let $\mathcal{T}_{\mathcal{D}}$ be tasks that require knowledge from $\mathcal{D}$ to answer correctly, such as question answering. 2) \textbf{Capability Preservation}: Let $\mathcal{T}_{\text{gen}}$ be a set of tasks that test for general capabilities already present in $M_{\text{post}}$, such as instruction following, math reasoning or coding.

We report the trained model $M_{\text{new}}$'s performance on both these task suites, denoted by $\text{Acc}(M_{\text{new}}; \mathcal{T}_{\mathcal{D}})$ and $\text{Acc}(M_{\text{new}}; \mathcal{T}_{\text{gen}})$ respectively. A successful adaptation procedure would yield an $M_{\text{new}}$ that maximizes the adaptation performance $\text{Acc}(M_{\text{new}}; \mathcal{T}_{\mathcal{D}})$ while minimizing the drop in prior capabilities $\text{Acc}(M_\text{post}; \mathcal{T}_{\text{gen}}) - \text{Acc}(M_\text{new}; \mathcal{T}_{\text{gen}})$.

\subsection{Standard Finetuning as Likelihood Optimization}
\label{sec:prelim}

Standard fine-tuning with the next-token prediction objective has been commonly used to adapt trained language models to new document corpora \citep{yang2024syntheticcontinuedpretraining, domainllama, domain_ft}. \vspace{-1mm}
\begin{equation}
\mathcal{L}_{\text{FT}} \;=\; \mathbb{E}_{d\sim\mathcal{D}} \left[ \sum_{t=1}^{|d|} -\log P(d_t \mid d_{<t}) \right] \vspace{-1mm}
\end{equation}
This directly encourages the model to fit the target distribution. In order to preserve capabilities of the initial model, most continual learning approaches modify this core objective to include KL \citep{ouyang2022training, deepseek} or weight regularization \citep{gu-etal-2024-model}, use low learning rates \citep{talr}, update smaller subspaces \citep{hu2022lora}, or augment $\mathcal{D}$ with replay data of demonstrations of capabilities that we want to preserve \citep{replay}.

However, these additional constraints do not fundamentally alter the learning objective. In fact, we find that they incur similar drops in performance on previously learned post-training capabilities as regular fine-tuning (\S\ref{sec:ftresults}). In this paper, we propose \textbf{Di}stillation via \textbf{S}plit \textbf{C}ontexts, a training method that overcomes these shortcomings.



\section{\methodname: Distillation via Split Contexts}
\label{sec:method} 

\subsection{Motivation}
\label{sec:intuition}
We hypothesize that the baseline methods we considered in Section \ref{sec:prelim} fail to preserve model performance due to their use of the next-token prediction loss as the primary task objective. Recent work~\citep{shenfeld2025rlsrazoronlinereinforcement} reports that fine-tuning leads to similar catastrophic forgetting when used for capability adaptation, such as verifiable task domains of math reasoning or scientific QA. A key insight from this work was that RL-based adaptation training mitigates forgetting because it favors solutions that minimize the KL distance between the updated and the initial policy. 

Note that a similar RL approach is unsuitable for knowledge adaptation of ``raw'' documents as a well-defined reward signal is lacking. On the other hand, adding KL regularization to standard fine-tuning does not reliably mitigate forgetting during knowledge adaptation across model sizes~\cite{talr}. \textbf{In \methodname, our key idea is to re-frame the knowledge adaptation task as a variant of context distillation, and use KL divergence as the main learning objective. }


First, we explain our intuition using QA as the inference-time task. Consider using model $M$ to generate responses for a target question $q$ under two inference settings. In the first case, we perform \textit{grounded} inference where generation is conditioned on a grounding document $d$ that contains the relevant information about the question $q$. Thus, we sample from $P_{M}(\cdot | d, q)$. In the second case, we remove the grounding document and simply sample from $P_{M}(\cdot | q)$. 

Intuitively, if $M$ already contains the knowledge from $d$ needed to answer $q$ in its parameters, we expect that the distributions $P_{M}(\cdot | d, q)$ and $P_{M}(\cdot | q)$ will be close to each other. Conversely, if $M$ does not contain the relevant knowledge, we can encourage it to internalize question-specific knowledge from the document $d$ by minimizing the KL divergence between the following two distributions: 
\begin{equation}
 M_\text{new} = \arg \min_{\hat{M} \in \mathcal{M}} \; \mathrm{KL} \left[ P_{M}(y \mid d, q) \;\middle\|\; P_{\hat{M}}(y \mid q) \right]
\label{eq:intuition} 
\end{equation}

This intuition forms the basis for \methodname. Note that the above objective is similar to context distillation in earlier works \cite{snell2022learning,askell2021general}. However, we cannot directly use this objective as our continual learning setup does not provide any $(d, q)$ pairs, and instead only provides raw text documents $d$ for training. 

In our approach, we do not attempt to generate an exhaustive set of questions that cover all knowledge in this corpus in order to operationalize the above intuition. Instead, we train $M$ to internalize document prefixes, by minimizing the KL divergence between its distributions over a suffix with and without conditioning on the prefix. 

Importantly, unlike most works involving distillation \citep{knowledge_distill, deepseek}, \textbf{our ``teacher'' and ``student'' models originate from \emph{the same policy}. Therefore, rather than deriving a signal from the distribution of a superior model, we derive our signal from the \emph{same} model with \emph{richer contextual conditioning}.} 


\begin{figure}[t]
\centering        \includegraphics[width=0.48\textwidth]{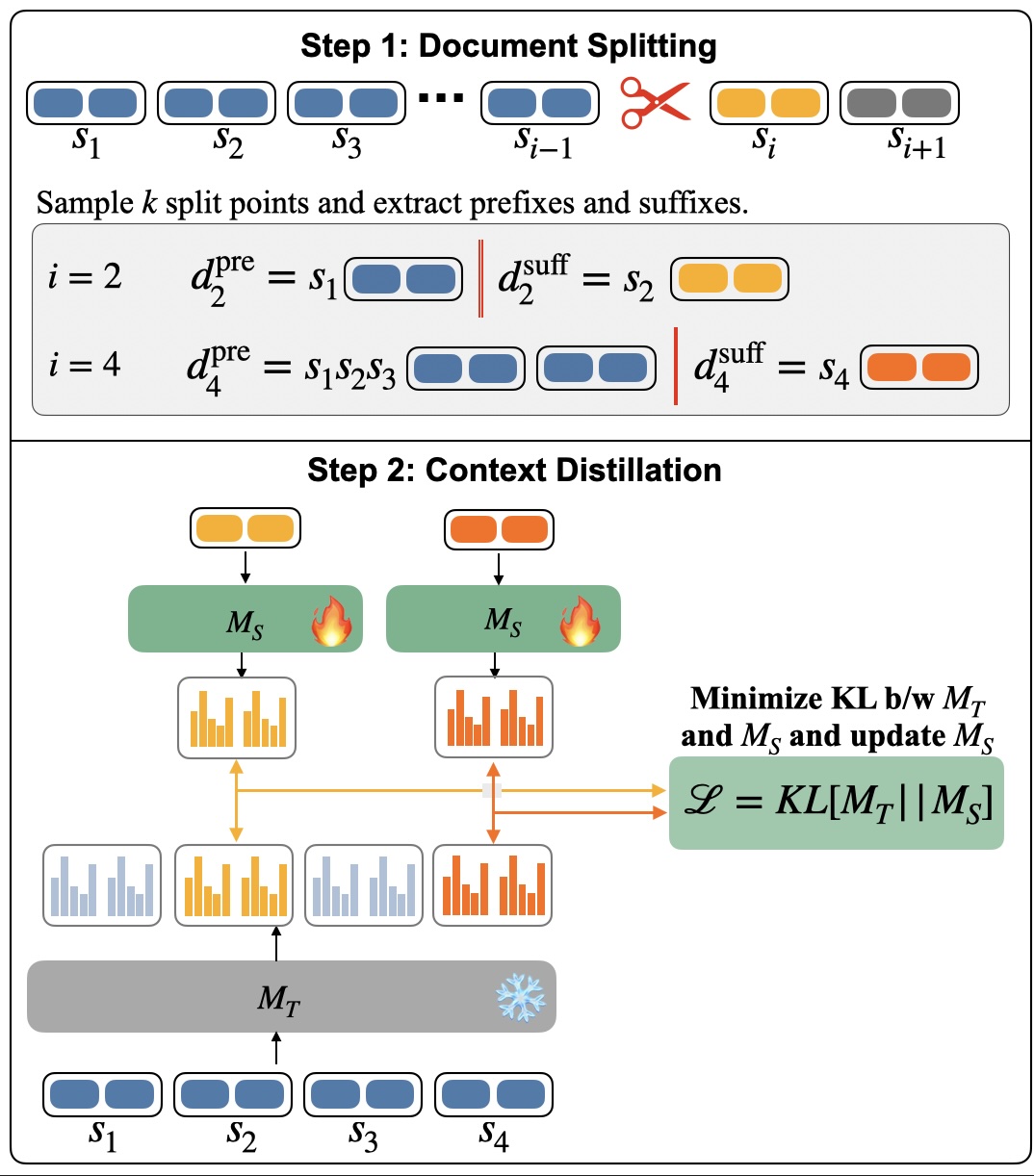} 
    \caption{Distillation via Split Contexts. First, we sample multiple split points and extract corresponding prefixes and suffixes. Then, we minimize the KL over the suffix distributions with context conditioning (teacher model $M_T$) and without (student model $M_S$). Our training only updates the student model $M_S$.}
    \label{fig:method} \vspace{-5mm}
\end{figure}

\subsection{\methodname~Algorithm}

Figure \ref{fig:method} outlines our approach. Let $d = (s_1, s_2, \cdots s_n)$ denote the sentences in a document $d \in \mathcal{D}$. We initialize a model $M_T$ as a frozen copy of the initial post-trained model $M_\text{post}$; this will serve as the teacher model and its weights will not be updated during training. Separately, we initialize the trainable student model $M_S$ with the parameters of $M_\text{post}$.


Below, we elaborate on the learning objective for each document $d$ in our training loop: 

\textbf{Step 1: Document Splitting}: For each document $d$ in the batch, we get $k$ split points at sentence boundaries; split point $i=2$ refers to the boundary between sentences $s_1$ and $s_2$. 
Specifically, we sample $k-1$ split points uniformly from $\{2,\cdots (n-1)\}$ and include $n$ as the last split point. 

Let $I$ refer to this set of split points. For each split point $i$, we extract a document prefix and a one-sentence suffix from $d$. Specifically, we set the prefix to be all preceding sentences and the suffix to be $s_i$:
\begin{equation*}
    d^\text{pre}_{i} = s_1s_2\cdots s_{i-1} \;\;\; d^\text{suff}_{i} = s_i 
\end{equation*}


\textbf{Step 2: Context Distillation}: For each suffix $d^\text{suff}_{i}$, we compute the teacher model's distribution conditioned on the corresponding prefix, that is $P_{M_T}(d^\text{suff}_{i} \mid d^\text{pre}_{i})$. We also compute the student model's distribution for the same suffix, that is $P_{M_S}(d^\text{suff}_{i})$, without conditioning on the context. Our document-level learning objective, therefore, is: 
\begin{equation}
\mathcal{L}_{\text{\methodname}}
=
\frac{1}{|\mathcal{I}|}\sum_{i \in \mathcal{I}}
\mathrm{KL}\!\left[
P_{M_T} (d_i^\text{suff} \mid d_i^\text{pre})
\|
P_{M_S} (d_i^\text{suff}) 
\right]
\label{eq:disc} 
\end{equation}

Drawing an analogy with Eq.\ref{eq:intuition} and building on the intuition from \S\ref{sec:intuition}, we can interpret the above equation as encouraging the model to internalize
the knowledge in document prefixes by minimizing the KL between the suffix distributions with and without conditioning on the prefix.

We update the student model $M_S$ with this KL-based loss objective. At the end of training, our algorithm returns $M_\text{new}$ as the knowledge adapted model. \vspace{-2mm}

\subsection{Efficiency Discussion} 

Note that Eq.~\ref{eq:disc} involves computing KL for $|I|$ different suffixes. An inefficient implementation of the objective would require $2\times|I|$ forward passes for the student and the teacher models combined. However, we emphasize that we can efficiently extract these in a single forward pass of the teacher and student models each. 

As shown in Figure~\ref{fig:method}, we perform one single forward pass with the teacher model using regular attention, and straightforwardly extract the $|I|$ suffix distributions. For the student model, we can construct a new input by concatenating $d_i^{\text{suff}}$ for all $i \in I$ and removing attention connections in between different suffixes. This similarly allows for extracting the $|I|$ suffix distributions in one single forward pass. Therefore, the loss can be computed in two forward passes total. \vspace{-2mm}


\subsection{Connections to prior context distillation literature} 
\label{sec:conn-cd}
Under the broad umbrella of context distillation, prior works \cite{askell2021general,choi2022prompt,snell2022learning,padmanabhan2023propagating} differ not only in how contexts and suffixes are obtained but also the underlying goal of the learning algorithm. Table~\ref{tab:cd} outlines these differences. We characterize differences along three axes: the internalization target (e.g. new knowledge for \methodname), how the context to be internalized is obtained, and the loss tokens for which the KL divergence between the student and the teacher is minimized. As before, we use $d$ to denote a train datapoint. 

The first line of work aims to internalize a static context, such as a constitution \cite{askell2021general} or persona \cite{choi2022prompt}, for all training points. In constrast, \citet{snell2022learning} do not restrict to static contexts and extend context distillation to internalize reasoning processes of the teacher. Here, both the context and the loss tokens are generated by the teacher model. However, neither of these internalize the knowledge in the training datapoint $d$ itself. 

The work most closely related to ours is \citet{padmanabhan2023propagating} which also targets knowledge internalization. However, their method requires generating an explicit \textit{transfer set} of continuations to serve as the suffix, whereas \methodname~naturally extracts these from the given document itself. Consequently, \methodname~results in a more efficient algorithm by avoiding the expensive generation step during training.  \vspace{-2mm}

\begin{table}[]
    \centering
    \small
    \caption{Representative prior works in context distillation.}\vspace{-2mm}
    \setlength{\tabcolsep}{3pt}
    \begin{tabular}{l|c|c}
    \toprule
    \textbf{Internalization Target} & \textbf{Context} & \textbf{Loss Tokens} \\
    \midrule
     Constitution 
        & Static context $c$
        & \multirow{2}{*}{$d \in \mathcal{D}$} \\ 
       \cite{askell2021general} & for all $d \in \mathcal{D}$ & \vspace{1mm} \\
     Instructions + Reasoning
        & In-context exs \&
        & \multirow{2}{*}{$y \sim P_{M_T}(\cdot | d, c)$} \\
       \cite{snell2022learning} & $c \sim P_{M_T}(\cdot | d)$ &  \vspace{1mm} \\
    New Knowledge 
        & \multirow{2}{*}{$d \sim \mathcal{D}$} 
        & \multirow{2}{*}{$y \sim P_{M_T}(\cdot | d)$} \\
    \cite{padmanabhan2023propagating} & & \\
    \midrule
    New Knowledge (\methodname)
        & $d^{\text{pre}}_i \in d \sim \mathcal{D}$
        & $d^{\text{suff}}_{i} \in d \sim \mathcal{D}$ \\
    \bottomrule
    \end{tabular}
    \label{tab:cd} \vspace{-5mm}
\end{table}

\section{Experimental Setup}

\subsection{Evaluation Datasets and Benchmarks}

As discussed in \S\ref{sec:prb_def}, our goal is to internalize new knowledge found in a corpus $\mathcal{D}$ while retaining general capabilities of the post-trained model. Below, we describe the datasets $\mathcal{D}$ used in our experiments and their corresponding evaluation suites $\mathcal{T}_{\mathcal{D}}$. We also describe the benchmark tasks $\mathcal{T}_\text{gen}$ used to measure general capabilities.

\textbf{Evaluating Knowledge Adaptation.}
We use 2 document-length datasets $\mathcal{D}$ to continually train our post-trained LMs. These datasets are chosen to capture two orthogonal characteristics of update datasets -- new corpora that contradicts existing knowledge and domain-specialized knowledge.  \vspace{-2mm}

\begin{enumerate}[leftmargin=*, noitemsep,topsep=0pt] 
    \item \textbf{Knowledge Update Playground (KUP)} \cite{li2025memorization}, which consists of 5,000 synthetic news-style passages that reflect entity-specific changes in world knowledge. We use the multiple-choice QA dataset released with KUP (500 examples) to evaluate our trained models. These questions probe for memorization of new knowledge, particularly if models prioritize recent knowledge over outdated facts and other strong distractors.
    \item \textbf{Biomedical Dataset BioASQ} \cite{krithara2023bioasq}, a QA dataset with 3,226 documents and 209 associated expert-curated cloze-style questions. In order to evaluate our post-trained models on this dataset, we convert these questions into multiple-choice ones by generating strong distractors using GPT-5. The prompt is in Appendix \ref{app:bioasq-prompt}
\end{enumerate} \vspace{-3mm}

\textbf{Evaluating General Capabilities.} We measure catastrophic forgetting using a broad set of general capability tasks $\mathcal{T}_{\text{gen}}$. Specifically, we report results on established standard benchmarks Big-Bench Hard (BBH) \citep{bbh}, GPQA \citep{gpqa}, IFEval \citep{ifeval}, MATH-Hard \citep{hendrycks2021math}, MuSR \citep{musr}, MMLU-Pro \citep{mmlu-pro}, and the coding benchmark HumanEval \citep{chen2021codex}. These benchmarks probe orthogonal capabilities that are not present in the update dataset, allowing us to isolate catastrophic forgetting from overfitting or domain shift. 

\subsection{Baselines} 
We compare \methodname~against the following baselines: 

\textbf{1. Fine-tuning (FT) and variants.} Our primary baseline is standard fine-tuning with the next-token prediction objective. We also include baselines that augment this objective: \vspace{-1mm}

(i) \textbf{+KL regularization} which penalizes deviations from the base model's output distribution during fine-tuning. 

(ii) \textbf{+LoRA} \citep{hu2022lora}, which adapts the model using low-rank updates to a subset of weight matrices while keeping the base model frozen. This constrains parameter updates to a lower dimensional subspace, reducing the extent of model drift during FT. 

(iii) \textbf{+Rephrase}, which adapts the self-distillation approach introduced in \cite{yang2024self} for our setting. It aims to reduce off-policy supervision during finetuning by generating on-policy rephrases of each training document. In particular, we use the initial model $M_\text{post}$ itself to generate these document rephrases that are on-policy for $M_{\text{post}}$.

(iv) \textbf{Token-adaptive loss reweighting (TALR)} \citep{talr}, which dynamically adjusts the weight of each token's loss to the overall loss based on its predicted probability. Their strategy assigns lower weight to high loss tokens while ensuring that the weight is not concentrated on only a subset of tokens (refer to the original work for details). We follow the weighting curriculum prescribed by \cite{talr}.

\textbf{2. Context-Distillation baseline (CD base)} Of the prior works on context distillation, only \citet{padmanabhan2023propagating} is applicable to our setting. We describe their approach earlier in \S\ref{sec:conn-cd}. We tune the hyperparameters to our setting. More details in Appendix~\ref{app:training-details}.

\subsection{Models and Training Details}
We run experiments on four post-trained models to cover distinct model sizes and families: \qwenlarge, \qwensmall~\citep{qwen2.5}, \llama~\citep{llama3} and \qwenthree~\citep{yang2025qwen3}.

All models are trained in FP32 for one epoch. For KL regularization, we set $\beta = 0.1$, a relatively strong regularization value compared to those commonly used in prior work (e.g., \citealp{ouyang2022training}) in order to mitigate forgetting. For our LoRA variant, we apply LoRA with rank $r=16$ to the attention and MLP projection layers as per \citep{schulman2025lora}. When comparing baseline FT methods in \S\ref{sec:ftresults}, we use the same learning rate for all experiments for the same model; we select this as the LR that reports the highest adaptation performance on each respective dataset. We used this strategy as we were unable to run hyperparameter sweeps for each dataset, baseline method, and model combination. We conduct a wider learning rate sweep for standard finetuning, CD base and \methodname~in \S\ref{sec:disc_results}, to balance both adaptation performance and capability retention. See Appendix~\ref{app:training-details} for details. 

For evaluation, we use the LM Eval Harness \cite{eval-harness} to benchmark performance on general capabilities, and custom evaluation scripts for the adaptation tasks. As is standard, we use the respective model's chat template for all evaluations. To reduce variance in domain evaluations, we sample 5 answers for each question and use majority vote. \vspace{-3.5mm}

\begin{table*}[t]
\centering
\scriptsize

\setlength{\tabcolsep}{3pt}
\caption{Performance of finetuning baselines and its variants on two adaptation tasks: KUP and BioASQ. We find that standard finetuning generally improves adaptation task performance, but at the cost of substantial degradation of $M_\text{post}$'s IFEval, Math and Code skills. } \vspace{-2mm}
\resizebox{0.98\textwidth}{!}{
\begin{tabular}{lcccccccc|cccccccc}
\toprule

& \multicolumn{8}{c|}{\textsc{\textbf{KUP}}} & \multicolumn{8}{c}{\textsc{\textbf{BioASQ}}} \\
\cmidrule(lr){2-9}\cmidrule(lr){10-17}
Method
& BBH & GPQA & MMLU-P & MuSR & IFEval & Math & Code & KUP
& BBH & GPQA & MMLU-P & MuSR & IFEval & Math & Code & BioASQ \\
\midrule
\multicolumn{17}{l}{\hspace{5mm}\textbf{\qwenlarge}} \\ \midrule
$M_\text{post}$
& 53.69 & 30.29 & 42.89 & 39.81 & 80.70 & 49.85 & 78.66 & 15.46
& 53.69 & 30.29 & 42.89 & 39.81 & 80.70 & 49.85 & 78.66 & 70.19 \\
\midrule
FT
& \cellcolor{red!2}52.25 & \cellcolor{red!2}28.44 & \cellcolor{red!5}38.33 & \cellcolor{green!2}40.74
& \cellcolor{red!30}65.83 & \cellcolor{red!34}24.47 & \cellcolor{red!19}69.51 & \cellcolor{green!34}40.90
& \cellcolor{green!1}54.07 & \cellcolor{green!1}30.87 & \cellcolor{red!6}37.86 & \cellcolor{green!3}43.25
& \cellcolor{red!7}75.18 & \cellcolor{red!30}35.05 & \cellcolor{red!5}73.78 & \cellcolor{green!7}77.40 \\
+Rephrase
& \cellcolor{red!2}51.54 & \cellcolor{red!2}28.86 & \cellcolor{red!7}37.37 & \cellcolor{green!4}43.92
& \cellcolor{red!31}63.91 & \cellcolor{red!33}28.25 & \cellcolor{red!12}71.34 & \cellcolor{green!34}42.90
& \cellcolor{red!2}52.47 & \cellcolor{red!2}29.03 & \cellcolor{red!7}37.38 & \cellcolor{red!1}39.68
& \cellcolor{red!20}71.22 & \cellcolor{red!31}32.85 & \cellcolor{red!10}71.95 & \cellcolor{green!5}76.44 \\
+TALR
& \cellcolor{red!2}52.70 & \cellcolor{red!2}28.86 & \cellcolor{red!4}38.90 & \cellcolor{green!5}44.18
& \cellcolor{red!30}65.83 & \cellcolor{red!34}26.51 & \cellcolor{red!3}75.61 & \cellcolor{green!20}36.90
& 54.05 & \cellcolor{red!2}29.28 & 41.30 & 39.15
& 77.70 & 46.98 & 81.1 & \cellcolor{green!10}78.37 \\
+KL
& \cellcolor{red!1}53.39 & \cellcolor{red!1}30.03 & \cellcolor{red!5}38.28 & \cellcolor{green!3}42.99
& \cellcolor{red!25}69.78 & \cellcolor{red!34}27.57 & \cellcolor{red!4}75.00 & \cellcolor{green!34}40.10
& \cellcolor{red!21}53.81 & 29.95 & 42.50 & 49.40
& 77.94 & 41.27 & 82.32 & \cellcolor{green!10}78.37 \\
+LoRA
& \cellcolor{red!1}52.65 & \cellcolor{red!2}29.11 & \cellcolor{red!5}39.87 & 42.92 & \cellcolor{red!8}75.90 & \cellcolor{red!2}43.52 & 78.66 & \cellcolor{green!15}35.6 
& \cellcolor{red!6}48.78 & \cellcolor{green!1}30.37 &  \cellcolor{red!3}39.34 & \cellcolor{green!2}40.61 & \cellcolor{red!9}74.46 & \cellcolor{red!25}38.52 & \cellcolor{red!3}75.61 &  \cellcolor{green!9}76.44  \\ \midrule

\multicolumn{17}{l}{\hspace{5mm}\textbf{\llama}} \\ \midrule
$M_\text{post}$
& 50.63 & 29.70 & 37.68 & 38.76 & 85.25 & 20.92 & 70.12 & 20.10
& 50.63 & 29.70 & 37.68 & 38.76 & 85.25 & 20.92 & 70.12 & 79.33 \\
\midrule
FT
& \cellcolor{red!2}49.02 & \cellcolor{red!2}28.36 & \cellcolor{red!3}34.60 & \cellcolor{green!1}39.42 & \cellcolor{red!31}68.11 & \cellcolor{red!9}14.73 & \cellcolor{red!17}61.59 & \cellcolor{green!30}35.10
& 50.15 & 30.54 & 37.53 & 37.83
& 85.25 & \cellcolor{red!10} 16.99 & 64.02 & 80.77 \\
+Rephrase
& \cellcolor{red!2}48.72 & \cellcolor{red!3}27.10 & \cellcolor{red!4}33.49 & \cellcolor{green!2}39.81 & \cellcolor{red!33}64.99 & \cellcolor{red!16}12.61 & \cellcolor{red!28}57.32 & \cellcolor{green!32}37.40
& \cellcolor{red!3}47.06 & \cellcolor{red!3}26.93 & \cellcolor{red!11}30.73 & \cellcolor{green!2}39.95 & \cellcolor{red!22}75.30 & \cellcolor{red!21}11.40 & \cellcolor{red!31}54.27 & \cellcolor{red!34}78.85 \\
+TALR
& \cellcolor{red!1}50.58 & \cellcolor{red!2}27.52 & \cellcolor{red!2}35.40 & \cellcolor{red!1}38.36 & \cellcolor{red!26}73.62 & \cellcolor{red!9}14.73 &\cellcolor{red!12}62.80 & \cellcolor{green!27}32.40
& \cellcolor{red!2}48.98 & \cellcolor{red!1}29.53 &\cellcolor{red!7}31.97 & \cellcolor{green!3}41.40 & \cellcolor{red!4}81.18 & \cellcolor{red!14}13.29 &  \cellcolor{red!7}64.63 & \cellcolor{red!10}82.21 \\
+KL
& \cellcolor{red!1}50.39 & \cellcolor{red!3}27.10 & \cellcolor{red!4}33.93 & 38.76 & \cellcolor{red!33}64.87 & \cellcolor{red!5}16.10 & \cellcolor{red!19}60.98 & \cellcolor{green!29}33.70
& \cellcolor{red!2}49.71 & \cellcolor{red!2}27.94 &\cellcolor{red!6}32.69 & \cellcolor{red!2}37.30 & \cellcolor{red!5}80.70 & \cellcolor{red!13}13.37 &  \cellcolor{red!19}60.98 & \cellcolor{red!14}81.25\\
+LoRA
& \cellcolor{green!1}50.82 & \cellcolor{red!2}28.69 &\cellcolor{red!1}37.28 & \cellcolor{red!2}37.43 & \cellcolor{red!5}80.82 & \cellcolor{red!2}19.26 & \cellcolor{red!3}67.07 & \cellcolor{green!25}31.00 
& \cellcolor{green!1}51.29 & 29.70 & \cellcolor{red!1}37.02 & \cellcolor{red!2}37.04 & \cellcolor{green!1}85.49 & \cellcolor{red!6}16.01 & \cellcolor{red!15}62.20 & \cellcolor{green!2}81.25 \\ \midrule

\multicolumn{17}{l}{\hspace{5mm}\textbf{\qwensmall}} \\ \midrule
$M_\text{post}$
& 46.54 & 28.52 &  32.75 & 39.42 & 72.66 &  37.24 & 71.14 & 26.41
& 46.54 & 28.52 & 32.75 & 39.42 &  72.66 &  37.24 &  71.14 & 67.79 \\
\midrule
FT
 & \cellcolor{red!2}45.74 & \cellcolor{green!2}29.53 &\cellcolor{green!1}32.79 & \cellcolor{green!2}41.80 &  \cellcolor{red!16}64.39 & \cellcolor{red!29}23.87 &  \cellcolor{red!30}56.10 & \cellcolor{green!32}44.57
 & \cellcolor{green!1}46.66 & \cellcolor{red!2}27.77 & \cellcolor{green!2}33.63 & \cellcolor{green!2}40.74 & \cellcolor{red!2}70.86 & \cellcolor{red!13}29.68 & \cellcolor{red!10}64.63 & \cellcolor{green!6}73.08 \\
+Rephrase
& \cellcolor{red!1}46.43 & 28.52 & \cellcolor{green!2}34.15 & \cellcolor{green!5}43.92 & \cellcolor{red!26}61.03 & \cellcolor{red!28}24.55 & \cellcolor{red!4}67.07 & \cellcolor{green!26}38.06
& \cellcolor{red!1}46.00 & \cellcolor{red!1}28.36 & \cellcolor{red!2}30.74 & \cellcolor{green!2}41.40 & \cellcolor{red!22}62.71 & \cellcolor{red!27}25.08 &\cellcolor{red!23}60.98 & \cellcolor{green!2}69.71 \\
+TALR
& \cellcolor{green!2}47.91 & \cellcolor{green!2}29.61 & \cellcolor{green!3}36.06 & \cellcolor{red!2}37.70 & \cellcolor{red!2}70.50 & \cellcolor{red!2}36.25 & \cellcolor{green!2}73.13 & \cellcolor{green!30}41.86
& \cellcolor{green!7}51.97 & \cellcolor{green!1}29.03 & \cellcolor{green!3}35.79 & \cellcolor{green!2}41.80 & \cellcolor{green!2}75.18 & \cellcolor{red!3}34.44 & \cellcolor{red!12}64.02 & \cellcolor{green!2}69.71 \\
+KL
& \cellcolor{green!2}47.82 & \cellcolor{green!1}28.61 &\cellcolor{green!2}35.22 & \cellcolor{green!1}39.81 & \cellcolor{red!3}70.02 & \cellcolor{red!2}36.18 &  \cellcolor{red!2}69.51 & \cellcolor{green!25}37.35
& \cellcolor{red!2}45.24 & \cellcolor{green!2}29.78 &\cellcolor{green!2}33.69 & \cellcolor{green!2}40.34 & \cellcolor{red!7}67.03 & \cellcolor{red!19}28.17 &  \cellcolor{red!8}65.24 & 67.79\\
+LoRA
& \cellcolor{green!2}47.86 & \cellcolor{red!1}27.94 &\cellcolor{green!3}35.85 & \cellcolor{green!2}41.40 & \cellcolor{red!4}68.47 & \cellcolor{red!2}36.33 & \cellcolor{red!2}69.51 & \cellcolor{green!32}45.00
& \cellcolor{red!3}43.46 & \cellcolor{red!1}27.94 &\cellcolor{red!2}30.30 & \cellcolor{red!1}39.15 & \cellcolor{red!2}70.86 & \cellcolor{red!3}33.76 & \cellcolor{red!2}70.12 & \cellcolor{red!6}62.50  \\ 
\midrule
\multicolumn{17}{l}{\hspace{5mm}\textbf{\qwenthree}} \\ \midrule
$M_\text{post}$
& 47.37 & 25.34 & 34.73 & 40.48 & 85.49 & 79.68 & 84.15 & 16.30
& 47.37 & 25.34 & 34.73 & 40.48 & 85.49 & 79.68 & 84.15 & 75.48 \\
\midrule
FT
& \cellcolor{green!13}53.88 & \cellcolor{green!2}26.43 & \cellcolor{red!16}26.89 & \cellcolor{red!13}33.99
& \cellcolor{red!20}75.54 & \cellcolor{red!16}71.90 & \cellcolor{red!20}74.39 & \cellcolor{green!38}35.40
& \cellcolor{green!12}53.46 & \cellcolor{green!1}26.01 & \cellcolor{green!6}37.48 & \cellcolor{red!7}36.98
& \cellcolor{green!5}87.89 & \cellcolor{green!2}80.82 & \cellcolor{green!2}85.37 & \cellcolor{green!26}88.69 \\
+Rephrase
& \cellcolor{green!12}53.46 & \cellcolor{green!1}26.01 & \cellcolor{red!0}34.53 & \cellcolor{red!10}35.45 & \cellcolor{red!71}49.88
& \cellcolor{red!27}66.09 & \cellcolor{red!18}75.00 & \cellcolor{green!32}32.20
& \cellcolor{green!14}54.57 & \cellcolor{green!1}26.01 & \cellcolor{green!14}41.58 & \cellcolor{red!1}39.81
& \cellcolor{green!3}86.93 & \cellcolor{green!7}83.01 & \cellcolor{green!2}85.37 & \cellcolor{green!13}82.21 \\
+TALR
& \cellcolor{green!12}53.60 & \cellcolor{green!2}26.26 & \cellcolor{red!12}28.90 & \cellcolor{red!8}36.64
& \cellcolor{red!25}73.02 & \cellcolor{red!21}69.26 & \cellcolor{red!7}80.49 & \cellcolor{green!25}29.00
& \cellcolor{green!13}53.84 & \cellcolor{green!1}25.76 & \cellcolor{green!10}39.54 & \cellcolor{green!2}41.53
& \cellcolor{green!1}86.09 & \cellcolor{green!2}80.82 & \cellcolor{red!5}81.71 & \cellcolor{green!15}83.17 \\
+KL
& \cellcolor{green!12}53.62 & \cellcolor{green!2}26.09 & \cellcolor{red!14}27.89 & \cellcolor{red!11}34.79
& \cellcolor{red!15}77.82 & \cellcolor{red!19}70.32 & \cellcolor{red!9}79.88 & \cellcolor{green!29}30.80
& \cellcolor{green!8}51.61 & \cellcolor{green!1}26.01 & \cellcolor{green!11}40.36 & \cellcolor{green!0}40.61
& \cellcolor{green!4}87.41 & \cellcolor{green!4}81.65 & \cellcolor{red!2}82.93 & \cellcolor{green!19}85.10\\
+LoRA
& \cellcolor{green!10}52.63 & \cellcolor{green!1}26.01 & \cellcolor{green!12}40.67 & \cellcolor{red!8}36.38 & \cellcolor{green!5}88.01 & \cellcolor{green!3}81.27 & \cellcolor{green!4}85.98 & \cellcolor{green!27}29.90 
& \cellcolor{green!15}54.76 & \cellcolor{green!2}26.17 & \cellcolor{green!7}38.31 & \cellcolor{green!5}42.86 & \cellcolor{green!2}86.45 & \cellcolor{red!0}79.58 & \cellcolor{green!4}85.98 & \cellcolor{green!15}83.17 \\ 
\bottomrule
\end{tabular}
}
\label{tab:combined_ft}
\vspace{-4mm}
\end{table*}

\section{Results} 
First, we discuss the performance and limitations of standard fine-tuning and its variants in \S\ref{sec:ftresults}. Then, in \S\ref{sec:disc_results}, we show that \methodname~successfully mitigates forgetting during continual adaptation compared to baselines.


\subsection{Performance of Fine-Tuning and its Variants}
\label{sec:ftresults}
Table~\ref{tab:combined_ft} reports these results.


\textbf{Standard finetuning reliably reports high adaptation performance but at the cost of substantial degradation to general capabilities.} For example, for \qwenlarge, finetuning reports a large gain on KUP adataption (15.46 $\rightarrow$ 40.9), but this is accompanied by sharp drops in instruction following (IFEval:-14.9), mathematical reasoning (Math:-25.4) and coding abilities (Code:-9.15). We find similar trends for \llama~on KUP: FT increases KUP performance by 15 points, but reduces IFEval by -17.1, Math by -6.2 and Code by -8.5. For \qwenthree, FT increases KUP performance by 19 points at the cost of reducing IFEval by -10.0 and MATH by -7.80. 

While not directly comparable, we observe that finetuning on BioASQ generally leads to lesser catastrophic forgetting than KUP. We attribute this to the different levels of ``domain-shift'' that these datasets reflect. To elaborate, note that all $M_\text{post}$ models report very poor performance on KUP and high gains after continual adaptation. On the other hand, BioASQ reports relatively higher initial performance, and lesser improvement after finetuning.

\textbf{The highest degradation is generally observed in math, code and instruction-following capabilities.} Notably, MuSR, while also a reasoning task similar to math and code, does not suffer from catastrophic forgetting. We hypothesize that math, code and instruction-following correspond to capabilities that the model acquires during post-training, as opposed to pre-training, and therefore might be more brittle. We verify this hypothesis in \S\ref{sec:base_vs_post}. Interestingly, we observe that finetuning improves \qwenthree~performance on BBH for both domains by $\sim$6 points, and MMLU-P for BioASQ by $\sim$3-7 points. Our analysis in \S\ref{sec:base_vs_post} shows that these performance numbers after adaptation are similar to that of $M_\text{post}$'s pretrained checkpoint, providing further evidence that changes to the model after post-training might be brittle under further finetuning.


\begin{table*}[t]
\centering
\scriptsize
\caption{Comparing \methodname~to standard fine-tuning and CD base following adaptation on KUP and BioASQ. The superscript $^\text{CP}$ indicates model checkpoints achieving high domain adaptation while experiencing limited forgetting. Across datasets, \methodname~consistently reports a stronger trade-off between these two goals.} \vspace{-2mm}
\setlength{\tabcolsep}{3pt}
\label{tab:disc_results}
\resizebox{0.97\textwidth}{!}{
\begin{tabular}{l|cccccccc|cccccccc}
\toprule
& \multicolumn{8}{c}{\textsc{\textbf{KUP}}} & \multicolumn{8}{c}{\textsc{\textbf{BioASQ}}} \\
\cmidrule(lr){2-9}\cmidrule(lr){10-17}
\textbf{Method}
& BBH & GPQA & MMLU-P & MuSR & IFEval & Math & Code & KUP
& BBH & GPQA & MMLU-P & MuSR & IFEval & Math & Code & BioASQ \\
\midrule
\multicolumn{17}{l}{\hspace{5mm}\textbf{\qwenlarge}} \\
\midrule
$M_\text{post}$
& 53.69 & 30.29 & 42.89 & 39.81 & 80.70 & 49.85 & 78.66 & 15.46
& 53.69 & 30.29 & 42.89 & 39.81 & 80.70 & 49.85 & 78.66 & 70.19 \\
FT (from \S\ref{sec:ftresults})
& \cellcolor{red!2}52.25 & \cellcolor{red!2}28.44 & \cellcolor{red!5}38.33 & \cellcolor{green!2}40.74
& \cellcolor{red!30}65.83 & \cellcolor{red!34}24.47 & \cellcolor{red!19}69.51 & \cellcolor{green!34}40.90
& \cellcolor{green!1}54.07 & \cellcolor{green!1}30.87 & \cellcolor{red!6}37.86 & \cellcolor{green!3}43.25
& \cellcolor{red!7}75.18 & \cellcolor{red!30}35.05 & \cellcolor{red!5}73.78 & \cellcolor{green!3}77.40 \\ \midrule
FT$^\text{CP}$ 
& \cellcolor{red!2}53.58 & 31.80 & 42.01 & 41.14
& 76.74 & 45.24 & 81.71 & \cellcolor{green!20}36.3
& 52.66 & 29.78 & 42.67& 39.02
& 80.46 & 50.53 & 80.49 & \cellcolor{green!25}73.08 \\
CD base$^\text{CP}$ & 54.42 & 30.62 & 43.02 & 40.08
& 79.38 & 49.24 & 82.32 & 32.10
& 54.89 & 30.45 & 43.41 & 39.95
& 81.65 & 46.83 & 81.10 & \cellcolor{green!30}76.92 \\
DiSC$^\text{CP}$ 
& 54.07 & 28.27  & 41.27  & 40.48  & 77.70 & 44.34 & 76.22 & \cellcolor{green!45}44.18
& 53.81 & 28.86 & 42.10 & 41.27 & 80.82 & 46.75 & 76.83 & \cellcolor{green!45} 81.73 \\



\toprule
\multicolumn{17}{l}{\hspace{5mm}\textbf{\llama}} \\
\midrule
$M_\text{post}$
& 50.63 & 29.70 & 37.68 & 38.76 & 85.25 & 20.92 & 70.12 & 20.10
& 50.63 & 29.70 & 37.68 & 38.76 & 85.25 & 20.92 & 70.12 & 79.33 \\
FT (from \S\ref{sec:ftresults})
& \cellcolor{red!3}49.02 & \cellcolor{red!2}28.36 & \cellcolor{red!6}34.60 & \cellcolor{green!1}39.42 & \cellcolor{red!37}68.11 & \cellcolor{red!13}14.73 & \cellcolor{red!18}61.59 & \cellcolor{green!24}35.10
& 50.15 & 30.54 & 37.53 & 37.83
& 85.25 & \cellcolor{red!10} 16.99 & 64.02 & \cellcolor{green!5}80.77\\ \midrule
FT$^\text{CP}$
& \cellcolor{red!2}51.17 & 29.28 & 37.87 & 38.36
& 82.01 & 20.54 & 70.12 & 33.9 
& 50.15 & 30.54 & 37.53 & 37.83
& 85.25 & \cellcolor{red!10} 16.99 & 64.02 & \cellcolor{green!5}80.77 \\
CD base$^\text{CP}$ & 50.65 & 28.59 & 36.31 & 38.23 
& \cellcolor{red!10}79.74 & 18.66 & 74.39 & \cellcolor{green!24} 35.80
& 51.48 & 30.20 & 37.11 & 38.89
& 84.29 & 20.02 & 67.68 & \cellcolor{green!5}80.29 \\
DiSC$^\text{CP}$
& 51.73 & 29.95  & 36.34  & 42.06  & \cellcolor{red!10} 79.14 & 19.86 & 65.24 & \cellcolor{green!45}44.88
& 51.85 & 29.45 & 35.85 & 39.15 & 85.01 & 18.96 & 64.63 & \cellcolor{green!10}81.73 \\



\toprule
\multicolumn{17}{l}{\hspace{5mm}\textbf{\qwensmall}} \\
\midrule
$M_\text{post}$
& 46.54 & 28.52 &  32.75 & 39.42 & 72.66 &  37.24 & 71.14 & 26.41
& 46.54 & 28.52 & 32.75 & 39.42 &  72.66 &  37.24 &  71.14 & 67.79 \\
FT (from \S\ref{sec:ftresults})
 & \cellcolor{red!2}45.74 & \cellcolor{green!3}29.53 &32.79 & \cellcolor{green!7}41.80 &  \cellcolor{red!24}64.39 & \cellcolor{red!40}23.87 &  \cellcolor{red!45}56.1 & \cellcolor{green!45}44.57
 & 46.66 & \cellcolor{red!2}27.77 & \cellcolor{green!3}33.63 & \cellcolor{green!4}40.74 & \cellcolor{red!6}70.86 & \cellcolor{red!27}29.68 & \cellcolor{red!24}64.63 & \cellcolor{green!45}73.08 \\ \midrule
FT$^\text{CP}$
& 48.22 & 30.12 & 35.78 & 39.29 & 69.66 & 32.70 & 75.00 & \cellcolor{green!4}38.5
& 46.66 & 28.44 & 32.38 & 39.02 & 71.58 & 37.46 & 65.85 & 64.42 \\
CD Base$^\text{CP}$ & 47.82 & 29.11 & 34.78 & 41.27
& 71.58 & 38.07 & 67.68 & \cellcolor{green!30}41.5
& 48.10 & 29.45 & 36.32 & 40.74
& 70.74 & 35.95 & 71.34 & 65.38 \\
DiSC$^\text{CP}$
& 47.80 & 29.61 & 35.40 & 41.40 & 70.02 & 35.95 & 73.08 & \cellcolor{green!35}42.27
& 47.80 & 27.18 & 35.99 & 39.81 & 72.42 & 37.08 & 72.56 & \cellcolor{green!35} 71.15 \\
\midrule
\multicolumn{17}{l}{\hspace{5mm}\textbf{\qwenthree}} \\
\midrule
$M_\text{post}$
& 47.37 & 25.34 & 34.73 & 40.48 & 85.49 & 79.68 & 84.15 & 16.30
& 47.37 & 25.34 & 34.73 & 40.48 & 85.49 & 79.68 & 84.15 & 75.48 \\
FT (from \S\ref{sec:ftresults})
& \cellcolor{green!13}53.88 & \cellcolor{green!2}26.43 & \cellcolor{red!16}26.89 & \cellcolor{red!13}33.99
& \cellcolor{red!20}75.54 & \cellcolor{red!16}71.90 & \cellcolor{red!20}74.39 & \cellcolor{green!38}35.40
& \cellcolor{green!12}53.46 & \cellcolor{green!1}26.01 & \cellcolor{green!6}37.48 & \cellcolor{red!7}36.98
& \cellcolor{green!5}87.89 & \cellcolor{green!2}80.82 & \cellcolor{green!2}85.37 & \cellcolor{green!26}88.69 \\ \midrule
FT$^\text{CP}$ 
& \cellcolor{green!18}56.59 & \cellcolor{green!1}25.92 & \cellcolor{red!0}34.51 & \cellcolor{red!6}37.30 & \cellcolor{red!4}83.69 & \cellcolor{green!4}81.80 & \cellcolor{red!7}80.49 & \cellcolor{green!33}32.60
& \cellcolor{green!12}53.46 & \cellcolor{green!1}26.01 & \cellcolor{green!6}37.48 & \cellcolor{red!7}36.98
& \cellcolor{green!5}87.89 & \cellcolor{green!2}80.82 & \cellcolor{green!2}85.37 & \cellcolor{green!26}88.69 \\
CD base$^\text{CP}$ 
& \cellcolor{green!15}54.80 & \cellcolor{green!1}26.01 & \cellcolor{green!9}39.20 & \cellcolor{green!2}41.40
& \cellcolor{red!19}75.90 & \cellcolor{green!2}80.91 & \cellcolor{red!1}83.54 & \cellcolor{green!24}28.30
& \cellcolor{green!14}54.39 & \cellcolor{green!1}26.01 & \cellcolor{green!11}40.30 & \cellcolor{green!2}41.27
& \cellcolor{red!0}85.49 & \cellcolor{green!4}81.72 & \cellcolor{red!4}82.32 & \cellcolor{green!19}85.10 \\
DiSC$^\text{CP}$ 
& \cellcolor{green!17}55.88 & \cellcolor{green!2}26.17 & \cellcolor{green!10}39.65 & \cellcolor{red!2}39.29 & \cellcolor{red!5}83.21 & \cellcolor{red!4}77.79 & \cellcolor{red!4}82.32 & \cellcolor{green!42}37.40
& \cellcolor{green!13}53.98 & \cellcolor{green!1}26.01 & \cellcolor{green!16}42.54 & \cellcolor{green!3}41.80 & \cellcolor{red!3}83.93 & \cellcolor{green!1}80.21 & \cellcolor{red!0}84.15 & \cellcolor{green!29}89.90 \\ \bottomrule
\end{tabular}
}

\end{table*}

\textbf{All fine-tuning variants trade off adaptation or capability preservation at the cost of the other without universal improvements.} Both TALR and KL regularization mitigate forgetting in select settings, but these gains are inconsistent across models and datasets. For \qwensmall~trained on KUP, both approaches eliminate forgetting on IFEval, Math and Code compared to finetuning, at the cost of slightly worse domain adaptation. However, on BioASQ, TALR and KL fail to retain their Code and Math capabilities respectively. 
Added to this, these approaches also fail to mitigate forgetting for \qwenlarge, \qwenthree, and \llama.

Rephrase shows similarly unreliable improvements. For both \qwenlarge~and \llama, FT + Rephrase is the strongest adaptation baseline for KUP (improves over regular finetuning by 2+ points), but causes similar (or higher) drops in Math, IFEval and Code. Moreover, it reports a much lower adaptation result compared to finetuning on BioASQ. 


Finally, while LoRA does reduce forgetting compared to standard finetuning (e.g. only -4.80 on IFEval compared to -14.87 w/ finetuning for \qwenlarge), it is most often accompanied by lower domain performance (e.g., 35.6 vs 40.9 for the same model) on KUP. 

Overall, these results highlight that none of the variants of finetuning are able to consistently mitigate forgetting while maintaining high knowledge adaptation performance. 


\subsection{Context-distillation methods v/s regular fine-tuning}
\label{sec:disc_results} 
Next, we study the continual adaptation performance of context-distillation methods, including \methodname. We choose regular FT as the representative method from \S\ref{sec:ftresults}. 

We report results in Table~\ref{tab:disc_results} for the checkpoint that achieves the highest adaptation score while experiencing limited forgetting when varying learning rates. We allow a maximum of $\sim$5 point drop on each of IFEval, Math and HumanEval. 
For an apples-to-apples comparison, we use this same selection criterion for all methods. We call these the \textbf{capability-preserving} checkpoints, and use the superscript $^\text{CP}$ to highlight these in the table.  In Table~\ref{tab:max_disc_results} (Appendix \ref{app:max_kup_performance}), we also report results using the LR that maximizes adaptation performance on KUP for DiSC, as in \S\ref{sec:ftresults}. 

\textbf{\methodname~achieves the strongest adaptation gains with minimal forgetting.} Across all models and datasets, \methodname~reports the highest adaptation performance among the capability preserving checkpoints, and even outperforms unconstrained finetuning on domain adaptation in six out of eight settings. For example, on KUP, \llama, \qwenlarge~and \qwenthree~outperform FT$^\text{CP}$ and CD base$^\text{CP}$ by a massive 5-12 points, while \qwensmall~outperforms these baselines by 0.6 and 3.6 points respectively. These superior domain adaptation performances are accompanied by on par or better capability preserving performance. For KUP, the 5-12 point improvement in domain adaptation is accompanied by a mean 1.33 point drop in IFEval and Math performances compared to FT$^\text{CP}$. For Code, comparing FT$^\text{CP}$ and \methodname, we see $\sim$2 point increase for \qwenthree~ (where \methodname$^\text{CP}$'s adaptation is also $\sim$5 points better) and $\sim$5 point decrease for \llama~ and \qwenlarge~ (where \methodname$^\text{CP}$'s adaptation is $8-12$ points better). Overall, our results provide strong evidence that \textbf{\methodname~reports a much stronger trade-off between domain adaptation and forgetting compared to baselines. }

We see similarly strong results on BioASQ. Here, \methodname$^\text{CP}$ outperforms the adaptation performance of both baselines for all models; an $\sim$5 points improvement for \qwenlarge~and \qwensmall, and between 1-5 points improvement for the other two. These are accompanied by similar trade-offs with capability preserving performances as the KUP dataset.



\begin{figure}
    \centering
    \includegraphics[width=0.8\columnwidth]{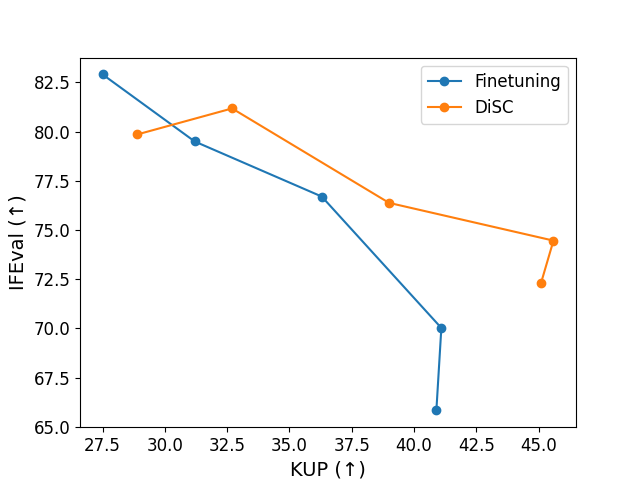}  \vspace{-2mm}
    \caption{IFEval vs KUP performances at different learning rates on \qwenlarge \text{ } for standard finetuning and \methodname.}
    \label{fig:lr_qwen2.5_7b} \vspace{-3mm}
\end{figure}

\textbf{Baseline Context Distillation reduces forgetting at the cost of inconsistent domain adaptation.} Baseline Context Distillation (CD base$^\text{CP}$) sometimes improves over FT$^\text{CP}$ in terms of domain adaptation, beating it in four of eight settings. However, its in-domain performance inconsistently competes with unconstrained finetuning and DiSC. While it matches these methods for \llama~on BioASQ, it dramatically underperforms both for \qwenlarge~ and \qwenthree ~on KUP.  This indicates that, although CD base is a reasonable strategy to mitigate forgetting, it is not well suited to achieve strong domain adaptation gains.

\paragraph{\methodname~is more robust to variance in learning rates compared to regular finetuning.} Table~\ref{tab:disc_results} only reports results for the LR that produced the capability preserving checkpoint for each baseline.  To further investigate how choice of learning rates impacts trade-offs between adaptation and preservation, we  report performance of DiSC and FT trained on KUP for five learning rates (LRs) between 1e-6 and 1e-5 in Fig~\ref{fig:lr_qwen2.5_7b}. We observe that as the LR increases, FT exhibits sharp degradation on IFEval, whereas DiSC sees a comparatively more gradual decline. At the highest LR, DiSC incurs roughly half of the forgetting on IFEval as FT (-8.4 vs -14.87) while achieving higher KUP performance. This also indicates that method is less sensitive to small variances in learning rate. 


\begin{table}[t]
\centering
\caption{Comparing $M_\text{base}$ and $M_\text{post}$ models across capabilities. Post-training primarily improves on IFEval, Math and Code.}\vspace{-2mm}
\setlength{\tabcolsep}{3pt}
\resizebox{0.9\columnwidth}{!}{
\begin{tabular}{l|cccccccc}
\toprule
Model
& BBH & GPQA & MMLU-P & MuSR & IFEval & Math & Code \\ \midrule
\multicolumn{8}{c}{\hspace{5mm}\textbf{\qwenlarge}} \\
\midrule
$M_\text{base}$ & 49.47 & 29.11 & 42.08 & 38.10 & 57.19 & 27.42 & 56.10 \\
$M_\text{post}$
& 53.69 & 30.29 & 42.89 & 39.81 & 80.70 & 49.85 & 78.66  \\ \midrule
\multicolumn{8}{c}{\hspace{5mm}\textbf{\llama}} \\
\midrule
 $M_\text{base}$
 & 46.02 & 31.54 & 32.39 & 37.17 & 15.59 & 5.06 &  43.29 \\
$M_\text{post}$
& 50.63 & 29.70 & 37.68 & 38.76 & 85.25 & 20.92 & 70.12 \\ \midrule
\multicolumn{8}{c}{\hspace{5mm}\textbf{\qwensmall}} \\ \midrule
Base
 & 43.55 & 26.93 &  32.39 & 40.08 & 45.32 &  16.62 & 36.59 \\
$M_\text{post}$
& 46.54 & 28.52 &  32.75 & 39.42 & 72.66 &  37.24 & 71.14 \\ \midrule
\multicolumn{8}{c}{\hspace{5mm}\textbf{\qwenthree}} \\ \midrule
Base
& 56.52 & 28.36 &  47.41 & 38.10 & 61.99 & 26.89 & 55.49\\
$M_\text{post}$
& 47.37 & 25.34 & 34.73 & 40.48 & 85.49 & 79.68 & 84.15  \\
\bottomrule
\end{tabular}
}
\label{tab:base_vs_post}
\end{table}

\section{Analysis}
\label{sec:analysis}
\subsection{Which Tasks does Finetuning Harm?}
\label{sec:base_vs_post} 
Our results in \S\ref{sec:ftresults} showed that IFEval, Math, and Code were the only tasks for which a consistent and substantial degradation in performance was observed after finetuning. We hypothesize that tasks that reflect skills acquired during post-training, i.e. those for which $M_{\text{post}}$ is significantly better compared to $M_{\text{base}}$, are ones that degrade the most during continual finetuning. To evaluate this, we report results for all four base and post-trained models. 

Results in Table~\ref{tab:base_vs_post} confirm our hypothesis. The only tasks which see a meaningful improvement after post-training across all four models were indeed IFEval, MATH, and HumanEval. Furthermore, we compute the Pearson correlation between post training gain: $\Delta_{\text{post}}(t)=\text{Instruct}(t)-\text{Base}(t)$ and continual FT change: $\Delta_{\text{FT}}(t)=\text{Instruct}(t) - \text{FT}(t)$ for all tasks $t$. Across all three models we observed a strong correlation between $\Delta_{\text{post}}(t)$ and $\Delta_{\text{FT}}(t)$: 0.83 for \qwenlarge, 0.98 for \llama, 0.94 for \qwensmall, and 0.56 for \qwenthree. Interestingly, this holds \emph{even for tasks for which the base model outperforms the post-trained version}. For instance, \textsc{Qwen3-8B-Base} outperforms \qwenthree \text{ }at BBH and MMLU-Pro, and in some cases finetuning on unrelated tasks \emph{improves} performance on these benchmarks (e.g., FT on KUP improves BBH 47.37 $\rightarrow$ 53.88). This suggests that standard finetuning systematically reverts model behavior towards its pretrained checkpoint. \vspace{-2mm}


\subsection{Does KL Predict Forgetting?} \vspace{-2mm}
\citet{shenfeld2025rlsrazoronlinereinforcement} found that for a simple neural network finetuned on an MNIST-esque task, the average per-token KL loss between the finetuned and initial model on the training data was the best predictor of forgetting on downstream tasks. In order to validate whether this observation extends to our more realistic setting, we measure the per-token KL loss between the initialization model and the finetuned model with each of the baselines we used. We then compute the correlation between KL loss and forgetting on IFEval and MATH for \qwenlarge~and \llama. We report results in Table~\ref{tab:kl_forgetting_correlation}.

\begin{table}[t]
\centering
\caption{
Per-dataset correlation between KL divergence (measured on the train domain)
and forgetting magnitude on held-out tasks.
} \vspace{-2mm}
\resizebox{0.95\columnwidth}{!}{
\begin{tabular}{l l cc}
\toprule
\textbf{Model} & \textbf{Domain} & \textbf{IFEval} & \textbf{MATH} \\
 &  & Pearson & Pearson  \\
\midrule
\multirow{2}{*}{\textbf{\llama}} & KUP & $+0.22$ & $+0.54$ \\
& BioASQ & $-0.54 $ & $+0.39$ \\
\midrule
\multirow{2}{*}{\textbf{\qwenlarge}} & KUP & $+0.51 $ & $+0.50 $ \\
 & BioASQ & $-0.55 $ & $-0.40$ \\
\bottomrule
\end{tabular}
}
\label{tab:kl_forgetting_correlation} \vspace{-4mm}
\end{table}
For both models on KUP, the average per-token KL loss on the training data exhibits a moderate positive correlation with forgetting on IFEval and MATH. In other words, a larger KL divergence came with more forgetting. However, for BioASQ, KL divergence often had a \emph{negative} correlation with forgetting on IFEval (with the exception of \llama~ on MATH). This challenges the findings from \citet{shenfeld2025rlsrazoronlinereinforcement} that average per-token KL loss is a strong predictor of forgetting. However, we note that our analysis is limited by small sample sizes ($n=6$ per model/task combination). Future work may be able to investigate this on a larger set. \vspace{-1mm}



\section{Related Work}
\textbf{Model Editing} work aims to update factual or behavioral knowledge in pretrained language models while minimizing unintended side effects. Early approaches such as ROME \citep{meng2022rome} and MEMIT \citep{meng2023memit} perform targeted weight edits to implant new facts but often induce degradation on unrelated capabilities \citep{gu-etal-2024-model}. Other approaches explore learning update rules directly (e.g., MEND \citep{mitchell2022mend}) or avoiding weight changes altogether via external memories \citep{mitchell2022serac}. 
In contrast to direct weight surgery, \citet{padmanabhan2023propagating} demonstrate that context-based distillation can effectively propagate injected knowledge. In contrast to these works, which study the injection of simple statements, our work focuses on a more realistic continual learning scenario where we update on LMs on \textit{documents} similar to a new corpora.

\textbf{Catastrophic Forgetting} has been extensively studied in continual learning for neural networks. Foundational methods include regularization-based approaches such as Elastic Weight Consolidation \citep{kirkpatrick2017ewc} and Synaptic Intelligence \citep{zenke2017si}, which constrain updates to parameters deemed important for previous tasks, and distillation-based methods like Learning without Forgetting \citep{li2017lwf}. Another widely used strategy is replay, where a subset of prior data is mixed into training to stabilize performance across tasks \citep{rebuffi2017icarl, lopezpaz2017gem}. More recent work highlights the importance of optimization choices and update structure, showing that adaptive optimizers can exacerbate forgetting \citep{hsu2019}, while sparsity or selective updates can significantly improve retention \citep{gu-etal-2024-model}. In the context of LLMs, recent approaches such as TALR \citep{talr} and on-policy finetuning \citep{chen2025retaining} aim to balance domain adaptation with general capability preservation. Our evaluation builds directly on this literature by systematically KL regularization, sparse updates, loss reweighting, data rephrasing, and optimizer choice. Our results highlight that distillation-based updates offer a strong and underexplored avenue in continual learning. 

\section{Conclusion} \vspace{-2mm}
We study continual knowledge adaptation for post-trained LMs, which requires  learning new knowledge while preserving prior capabilities. Through comprehensive experiments, we show fine-tuning and its variants induce significant forgetting. To address this, we develop a new context distillation-based method and demonstrate substantially stronger adaptation-retention trade-offs than finetuning baselines on three models and two adaptation domains.

\section*{Acknowledgement}  \vspace{-2mm}
We thank Eunsol Choi and the Cornell NLP group for helpful discussions and feedback. This project was partially supported by NSF grant IIS-2433072, and a gift from Google. We gratefully acknowledge use of the research computing resources of the Empire AI Consortium, Inc, with support from Empire State Development of the State of New York, the Simons Foundation, and the Secunda Family Foundation. Shankar Padmanabhan is gratefully supported by a NSF GFRP. 

\section*{Impact Statement}  \vspace{-2mm}
This paper presents work whose goal is to advance the field of Machine
Learning. There are many potential societal consequences of our work, none
which we feel must be specifically highlighted here.

\bibliography{icml2026}
\bibliographystyle{icml2026}

\newpage

\appendix
\section{DiSC vs FT: Maximum Domain Performance}
\label{app:max_kup_performance}
We report the highest recorded performance for DiSC and FT in Table~\ref{tab:max_disc_results}. DiSC is able to achieve significantly more domain adaptation while incurring substantially less forgetting on instruction following, math, and code. 
\begin{table*}[t]
\centering
\scriptsize
\setlength{\tabcolsep}{3pt}
\caption{Comparing maximum in-domain performance of \methodname~ with that of standard fine-tuning}
\begin{tabular}{lcccccccc|cccccccc}
\toprule
& \multicolumn{8}{c|}{\textsc{\textbf{Qwen2.5-7B-Instruct}}} & \multicolumn{8}{c}{\textsc{\textbf{LLaMA3.1-8B-Instruct}}} \\
\cmidrule(lr){2-9}\cmidrule(lr){10-17}
\textbf{Method}
& BBH & GPQA & MMLU-P & MuSR & IFEval & Math & Code & KUP
& BBH & GPQA & MMLU-P & MuSR & IFEval & Math & Code & KUP \\
\midrule
$M_\text{post}$
& 53.69 & 30.29 & 42.89 & 39.81 & 80.70 & 49.85 & 78.66 & 15.46
& 50.63 & 29.70 & 37.68 & 38.76 & 85.25 & 20.92 & 70.12 & 20.10 \\
\midrule
FT (from \S\ref{sec:ftresults})
& \cellcolor{red!2}52.25 & \cellcolor{red!3}28.44 & \cellcolor{red!8}38.33 & \cellcolor{green!1}40.74 & \cellcolor{red!26}65.83 & \cellcolor{red!45}24.47 & \cellcolor{red!16}69.51 & \cellcolor{green!32}40.9
& \cellcolor{red!3}49.02 & \cellcolor{red!2}28.36 & \cellcolor{red!6}34.60 & \cellcolor{green!1}39.42 & \cellcolor{red!37}68.11 & \cellcolor{red!13}14.73 & \cellcolor{red!18}61.59 & \cellcolor{green!24}35.10 \\ 
DiSC$^\text{Max KUP}$ 
& 53.17 & 28.36 & 41.45 & 39.68 & 74.46 & 44.03 & 77.44 & \cellcolor{green!40}47.59
& 50.96 & 29.61 & 35.46 & 40.74 & 75.06 & 17.90 & 64.63 & \cellcolor{green!40}49.50 \\
\bottomrule
\end{tabular}
\label{tab:max_disc_results}
\end{table*}

\section{Further Training Details}
\label{app:training-details}
Following existing work on knowledge editing \citep{meng2022rome}\citep{padmanabhan2023propagating} as well as recent recommendations for LLM finetuning \citep{small_batch}, we train with batch size 1 for all experiments. We experimented with larger batch sizes and the results did not change substantially relative to reported trends. 

For the main finetuning experiments, we separately tuned learning rate according to what optimized performance on each domain. On KUP, for \qwenlarge and \qwensmall this resulted in learning rates of 1e-5, for \llama this resulted in a learning rate of 4e-6, and for \qwenthree this resulted in a learning rate of 1.5e-5. For BioASQ, this resulted in a learning rate of 1e-5 for \qwensmall, 5e-6 for \qwenlarge and \qwenthree,a and 1e-6 for \llama. 

For the comparisons between DiSC and FT, we conducted a hyperparameter sweep by evaluating both methods on a set of 10 learning rates between 5e-7 and 2e-5: (5e-7, 8e-7, 1e-6, 2e-6, 3e-6, 4e-6, 5e-6, 8e-6, 1e-5, 2e-5). We use softmax temperature $T=2.0$ for distillation experiments. We experimented with the number of split points $|I|$ to use in DiSC and found $|I|=5$ achieved good performance empirically, with $|I| > 5$ leading to only marginal improvements at the cost of more compute. 

We use the AdamW \citep{loshchilov2019decoupled} optimizer for all experiments, with weight decay $0.01$, betas of $(0.9, 0.999)$ and epsilon of $1e-8$. 

For LoRA experiments, we use a rank of $r=16$ with a lora alpha of $32$, dropout of $0.1$, and no bias. We apply this to all linear layers in the transformer blocks. 

\section{Evaluation Details}
\label{app:eval-details}
We use the lm-evaluation-harness \citep{eval-harness} for general capability evaluations. We evaluated every method on each task using greedy decoding and with their native chat template, with few shot examples cast into a multiturn format. For Qwen3, we evaluate all tasks with thinking mode enabled with the exception of BBH and MMLU-Pro. We find that these tasks suffer significant performance decreases with thinking mode enabled (e.g, MMLU-Pro drops from 34.73  to 11.32). Models were cast to bf16 for evaluation. 

For KUP and BioASQ evaluations, we sampled five answers from the model per question at temperature $1.0$ and used majority vote to decide. We request structured outputs with \texttt{\textless think\textgreater} and\texttt{\textless answer\textgreater}  tags to simplify parsing; we parse answers even when formatting is violated.

\section{Prompts}
\subsection{KUP}
\label{app:kup-prompt}
We include the prompt used for KUP evaluation below.

\newpage
\begin{figure*}[t]
\centering
\begin{minipage}{0.98\textwidth}
    \begin{promptbox}{\text{KUP Evaluation Prompt}}
    \begin{Verbatim}[fontsize=\footnotesize, breaklines=true, breakanywhere=true, breaksymbolleft=]
"""You will answer a multiple-choice question about {entity}.
- Put your reasoning in: <think> ... </think>
- Put ONLY the final answer letter (A/B/C/D) in: <answer> ... </answer>
- Do not include any other text outside these tags.

{question}
"""
\end{Verbatim}
    \end{promptbox}
\end{minipage}
\end{figure*}

\subsection{BioASQ}
\label{app:bioasq-prompt}
We include the prompt used to generate distractor options for BioASQ below. We also include a few samples of options generated by GPT-5 (alongside one correct answer) below.  

\begin{figure*}[t]
\centering
\begin{minipage}{0.98\textwidth}
    \begin{promptbox}{\text{BioASQ Distractor Generator Prompt}}
    \begin{Verbatim}[fontsize=\footnotesize, breaklines=true, breakanywhere=true, breaksymbolleft=]
'''You are a careful biomedical distractor writer.
You will be given: (1) a question, (2) the correct answer, and (3) supporting context.
Return exactly THREE semantically plausible but wrong answer choices. The wrong answer choices should be sufficiently close to the correct
answer that a knowledgeable test taker might confuse them.
Rules:
- Make them concise noun phrases (a few words).
- Same answer type/category as the correct answer.
- Must be clearly contradicted by or NOT supported by the context.
- Avoid negations like "not", "none of the above", or humorous/absurd options.
- Do not repeat the correct answer with trivial rephrasing.
- Keep them distinct from each other.
STRICT OUTPUT (JSON only):
{
  "distractors": ["<D1>", "<D2>", "<D3>"]'''
}
\end{Verbatim}
    \end{promptbox}
\end{minipage}
\end{figure*}

\begin{figure*}[t]
\centering
\begin{minipage}{0.98\textwidth}
    \begin{promptbox}{BioASQ Questions and Answers Alongside GPT-5 generated options}
    \begin{Verbatim}[fontsize=\footnotesize, breaklines=true, breakanywhere=true, breaksymbolleft=]
Question:
How many selenoproteins are encoded in the human genome?

Options:
(A) 23
(B) 24
(C) 26 
(D) 25 (correct)


Question:
Which is the receptor for substrates of chaperone-mediated autophagy?

Options:
(A) HSPA8 / HSC70
(B) LAMP2A (correct)
(C) LAMP2B
(D) LAMP1


Question:
Which enzyme is involved in the maintenance of DNA (cytosine-5-)-methylation?

Options:
(A) Dnmt2
(B) Dnmt1 (correct)
(C) Dnmt3a
(D) Dnmt3b


Question:
Which pituitary adenoma is a common cause of infertility in women?

Options:
(A) Corticotroph adenoma
(B) Somatotroph adenoma
(C) Gonadotroph adenoma
(D) Prolactinoma (correct)

\end{Verbatim}
    \end{promptbox}
\end{minipage}
\end{figure*}

Finally, we also include the evaluation prompt for BioASQ:

\begin{figure*}[t]
\centering
\begin{minipage}{0.98\textwidth}
    \begin{promptbox}{BioASQ Evaluation Prompt}
    \begin{Verbatim}[fontsize=\footnotesize, breaklines=true, breakanywhere=true, breaksymbolleft=]
        "Answer the multiple-choice question by reasoning briefly in <think>...</think> and then giving ONLY the numeral of the correct option inside <answer>...</answer>.\n\n"
        f"Question:\n{question}\n\n"
        f"Options:\n{opts_str}\n\n"
        "Rules:\n"
        " - Put your final choice as a numeral only (e.g., 2) inside <answer> tags.\n"
        " - Do NOT repeat the option text inside <answer>.\n"
        " - Choose exactly one option.\n"
\end{Verbatim}
    \end{promptbox}
\end{minipage}
\end{figure*}

\end{document}